\documentclass[11pt]{article}
\usepackage[final]{acl}
\usepackage{times}
\usepackage{latexsym}
\usepackage[T1]{fontenc}
\usepackage[utf8]{inputenc}
\usepackage{microtype}
\usepackage{graphicx}
\usepackage{amsmath}
\usepackage{amssymb}
\usepackage{amsthm}
\usepackage{algorithm}
\usepackage{algorithmic}
\usepackage{booktabs}
\usepackage{multirow}
\usepackage{array}
\usepackage{tabularx}
\usepackage{stfloats}
\usepackage{placeins}

\newcolumntype{L}[1]{>{\raggedright\arraybackslash}p{#1}}
\newcolumntype{Y}{>{\raggedright\arraybackslash}X}
\definecolor{PositiveGreen}{RGB}{0,128,0}
\definecolor{NegativeRed}{RGB}{180,0,0}

\title{Adaptive Pullback: Reweighting Supervision at Penalty Jumps\\in On-Policy Distillation}
\author{
  \textbf{Guangyu Liu}\textsuperscript{\normalfont *1}\quad
  \textbf{Siyuan Gan}\textsuperscript{\normalfont *1}\quad
  \textbf{Shilong Zhou}\textsuperscript{\normalfont *1}\\
  \textbf{Linjian Meng}\textsuperscript{\normalfont 2}\quad
  \textbf{Boyan Wang}\textsuperscript{\normalfont\textdagger 1}\quad
  \textbf{Jing Huo}\textsuperscript{\normalfont\textdagger 1}\\
  {\normalfont\textsuperscript{1}State Key Laboratory of Novel Software Technology, Nanjing University, Nanjing, China}\\
  {\normalfont\textsuperscript{2}Shanghai Artificial Intelligence Laboratory, Shanghai, China}\\
  {\normalfont\{liuguangyu, gansiyuan, zsdragon\}@smail.nju.edu.cn}\\
  {\normalfont\{boyanwang, huojing\}@nju.edu.cn}
}

\begin{document}
\maketitle
\begingroup
\renewcommand{\thefootnote}{\fnsymbol{footnote}}
\footnotetext[1]{Equal contribution.}
\footnotetext[2]{Corresponding authors.}
\endgroup
\begin{abstract}
On-policy distillation (OPD) supervises student-generated responses using sampled-token student--teacher log-ratios.
A positive log-ratio penalizes a token that the student assigns higher probability to than the teacher, but its current value does not distinguish sustained high penalties from abrupt increases along a response.
We examine these local changes by comparing consecutive blocks of tokens within the same response.
Our diagnostic identifies penalty jumps, abrupt increases in average penalty that persist on held-out tokens.
Across these jumps, penalties rise while teacher confidence decreases.
We call this pattern penalty--confidence tension.
Boundaries into high-penalty blocks do not exhibit comparable confidence changes, showing that local changes provide information beyond the current penalty value.
Motivated by this observation, we propose \textbf{Adaptive Pullback OPD (AP-OPD)}.
At each detected jump, AP-OPD uses the preceding block as a local reference and uniformly scales all supervision coefficients in the following block so that its weighted average penalty matches the preceding block's original mean.
This reference determines the scaling factor without an additional strength constant.
AP-OPD improves performance on both mathematical reasoning and code generation.
\end{abstract}

\begin{figure*}[t]
    \centering
    \includegraphics[width=\textwidth]{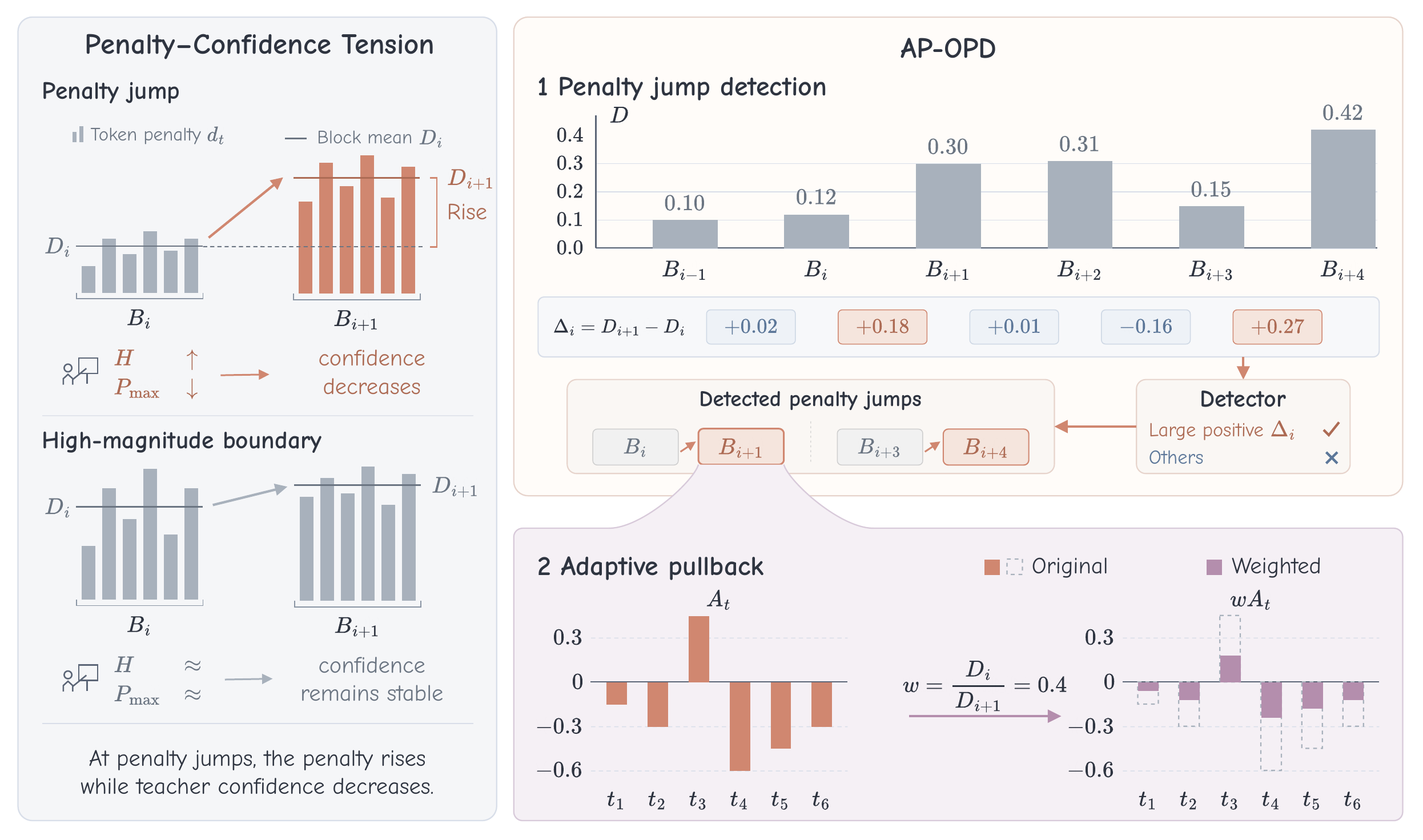}
    \caption{Penalty--confidence tension at penalty jumps: penalties rise while teacher confidence decreases.
    Adaptive Pullback uniformly scales post-jump supervision to match the preceding block's original mean penalty. Numbers are illustrative.
    $D_i$: block mean penalty; $H$: teacher entropy; $P_{\max}$: teacher top-1 probability; $w$: supervision weight.}
    \label{fig:overview}
    \vspace{-9pt} \end{figure*}

\section{Introduction}
On-policy distillation (OPD) transfers reasoning capabilities from a teacher model to a student through feedback on the student's own rollouts.
Training on student-generated prefixes brings the training distribution closer to the one visited at inference time and mitigates exposure bias
\citep{bengio2015scheduled,ranzato2016sequence,gu2024minillm,agarwal2024gkd,song2026opdsurvey}.
OPD also provides token-level teacher feedback throughout each response \citep{gu2024minillm,agarwal2024gkd}.
How to weight this supervision along a response is therefore an important question for OPD post-training.

In practical token-level OPD, supervision is determined by the sampled-token student--teacher log-ratio $r_t$ \citep{fu2026revisiting}.
A positive log-ratio penalizes a token that the student assigns higher probability to than the teacher.
The current penalty value describes how strongly a sampled token is penalized, but does not describe how that penalty changes along the response.
A high penalty can persist across neighboring segments or emerge as an abrupt increase.
Comparing consecutive blocks of tokens provides a local reference within the same response, allowing us to examine these increases separately from sustained high penalties.
We investigate whether local penalty changes carry information relevant to supervision allocation and whether using this information to reweight supervision improves training.

Our diagnostic analysis shows that local penalty changes reveal differences in teacher predictions that the current penalty value alone does not express.
We summarize each block by its block-averaged penalty, the mean positive log-ratio within a block.
We call an abrupt increase in the block-averaged penalty between consecutive blocks a penalty jump.
The analysis in Section~\ref{sec:diagnosis} shows that the increase persists on held-out tokens not used to detect the jump and is concentrated in the first block after it.
Across penalty jumps, the block-averaged penalty increases while teacher confidence decreases, as reflected in rising entropy and falling top-1 probability.
We call this observed pattern penalty--confidence tension.
Control boundaries entering blocks with high penalties do not exhibit comparable changes in teacher confidence.
This tension motivates examining whether supervision in the block after a jump should be reduced.

Motivated by penalty--confidence tension, we propose Adaptive Pullback OPD (AP-OPD), illustrated in Figure~\ref{fig:overview}.
The method uses the preceding block as a local reference and, at each detected jump, uniformly scales down all signed supervision coefficients in the following block so that its weighted block-averaged penalty matches the preceding block's original mean.
This rule uses an adjacent segment of the same response to set the reference.
A positive scaling factor preserves the signs and relative proportions of the token coefficients.
The reference determines the scaling factor without an additional strength constant.
AP-OPD detects jumps using log-probabilities already computed by OPD and requires no additional forward passes.

Training experiments support the value of this intervention.
AP-OPD improves performance, whereas a magnitude-cap rule that selects blocks by their penalty values and removes a similar amount of penalty brings no gain over vanilla OPD.
The removal amount alone therefore does not explain the improvement.
We also compare AP-OPD with FiRe-OPD, which weights supervision using teacher and student confidence \citep{li2026fireopd}, and IW-OPD, which uses accumulated prefix discrepancy \citep{xie2026iwopd}.
On mathematical reasoning, AP-OPD consistently improves macro-averaged Avg@$k$ over vanilla OPD under both jump detectors and obtains a higher macro average than the compared reweighting methods.
On code generation, with a different teacher and student, AP-OPD also achieves the highest macro average among the evaluated methods.

Our contributions are:
(1) We identify and diagnose penalty--confidence tension at penalty jumps in OPD, showing that local penalty changes provide information about changes in teacher predictions beyond the current penalty value.
(2) We propose Adaptive Pullback OPD, which uniformly scales supervision in the block after a jump using the preceding block as a local reference.
(3) We demonstrate gains on mathematical reasoning and code generation, and show that removing a comparable amount of penalty with a magnitude-cap rule does not yield the same benefit.

\section{Preliminaries}
\label{sec:prelim}

\subsection{Token-Level OPD}
Let $x$ be a prompt, $\pi_\theta$ the student, and $\pi_T$ a frozen teacher.
The student samples $\hat y_{1:T}\sim\pi_\theta(\cdot\mid x)$, and $c_t=(x,\hat y_{<t})$ denotes the context at position $t$.
With $\ell_t^S=\log\pi_\theta(\hat y_t\mid c_t)$ and $\ell_t^T=\log\pi_T(\hat y_t\mid c_t)$, the sampled log-ratio is
\begin{equation}
r_t=\ell_t^S-\ell_t^T .
\label{eq:log-ratio}
\end{equation}
The sequence-level reverse KL is $\mathbb E[\sum_t r_t]$ \citep{gu2024minillm}; its policy gradient uses suffix returns \citep{williams1992reinforce,sutton1999policy}.
Token-level OPD uses only the local term, with stop-gradient advantage $A_t=-\operatorname{sg}(r_t)$ \citep{fu2026revisiting}.
Averaging over the $N_{\mathrm{tok}}$ valid tokens in a mini-batch \citep{yu2025dapo} gives
\begin{equation}
\mathcal L
=-\frac{1}{N_{\mathrm{tok}}}\sum_{n}\sum_{t=1}^{T_n} w_{n,t}\,A_{n,t}\,\ell_{n,t}^S .
\label{eq:objective}
\end{equation}
Vanilla OPD sets $w_{n,t}=1$.
AP-OPD and the compared baselines all change the stop-gradient weights $w_{n,t}$.

\paragraph{Penalizing supervision.}
A single realization of $r_t$ can take either sign.
When $r_t>0$, the student assigns the sampled token a higher probability than the teacher, so $A_t<0$ and the update pushes that token down.
This case dominates in practice: in the vanilla OPD rollouts analyzed below, tokens with $r_t>0$ carry 83.5\% of $\sum_t|A_t|$.
We define the token penalty as
\begin{equation}
d_t=\operatorname{sg}\bigl([r_t]_+\bigr),\quad [u]_+=\max(u,0).
\label{eq:positive-divergence}
\end{equation}

\subsection{Block Profile and Jumps}
\label{sec:block-profile}
Single log-ratios fluctuate sharply between neighboring tokens, so we summarize a response at the level of contiguous blocks.
We partition its valid tokens into blocks $\mathcal B_1,\ldots,\mathcal B_M$ of $B$ tokens; only the last block may be shorter, and $n_i=|\mathcal B_i|$.
The block-averaged penalty $D_i$ is the mean positive log-ratio within a block, including zeros for nonpositive values; $s_i^2$ is the within-block sample variance of $d_t$ for $n_i\ge2$:
\begin{equation}
\scalebox{0.95}{$\displaystyle
D_i=\frac{1}{n_i}\sum_{t\in\mathcal B_i}d_t,\;
s_i^2=\frac{1}{n_i-1}\sum_{t\in\mathcal B_i}(d_t-D_i)^2 .
$}
\label{eq:block-profile}
\end{equation}
At boundary $i$, let $\Delta_i=D_{i+1}-D_i$.
A jump is a positive change judged sufficiently large by a chosen detector.
Let $\mathcal P$ denote the response's block penalty profile and required statistics, and $\operatorname{Det}(i;\mathcal P)\in\{0,1\}$ the detector's decision.
Over boundaries $i=1,\ldots,M-1$, the jump set is
\begin{equation}
\mathcal E_{\operatorname{Det}}=\bigl\{i:\Delta_i>0,\;
\operatorname{Det}(i;\mathcal P)=1\bigr\}.
\label{eq:jumps}
\end{equation}
The default $z$-gate accepts $i$ when $n_i,n_{i+1}\ge2$ and $\Delta_i>z\hat\sigma_i$, where
\begin{equation}
\hat\sigma_i=\sqrt{\frac{s_i^2}{n_i}+\frac{s_{i+1}^2}{n_{i+1}}} .
\label{eq:change}
\end{equation}
We fix $z=2$; because adjacent tokens are correlated, this is a standardized threshold, not a calibrated significance level.
For $i\in\mathcal E$, $\mathcal B_i$ and $\mathcal B_{i+1}$ are the pre- and post-jump blocks, and $\Delta_i$ is the rise.
Blocks only localize where to intervene; the loss in Eq.~\eqref{eq:objective} remains token-level.

\section{Diagnostic Analysis}
\label{sec:diagnosis}

\begin{figure}[t]
    \centering
    \includegraphics[width=\columnwidth]{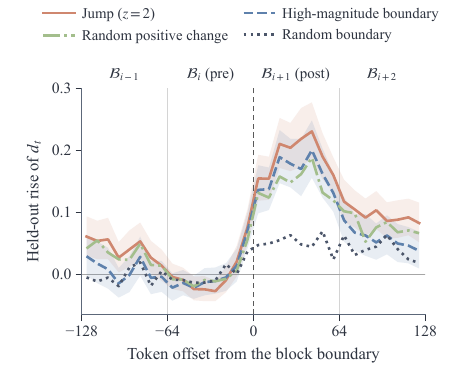}
    \caption{Held-out rise in token penalty $d_t$ around penalty jumps, abrupt increases in block-averaged penalty detected with the $z=2$ gate, and three control boundaries.
    The rise is relative to the mean penalty of the block before each boundary.
    Boundaries are selected on one half of each block's tokens and the curves are measured on the other half.
    The curves use 8-token bins and average both detection directions; bands for Jump and High-magnitude boundary are 95\% bootstrap intervals clustered by prompt.
    The boundary lies at offset 0, and $B=64$.}
    \label{fig:heldout}
    \vspace{-9pt} \end{figure}

We examine whether local changes in block-averaged penalty reveal changes in teacher predictions beyond the current penalty value.
We analyze 512 rollouts saved during vanilla OPD training at steps 70, 80, 90, and 100, one per distinct prompt, with the mathematical-reasoning student, teacher, and data of \S\ref{sec:setup}.
Jumps are detected with the default $z$-gate at $B=64$ and $z=2$, about twice per response.
We compare each jump with control boundaries from the same response at a similar relative position: the high-magnitude boundary, which enters the block with the highest block-averaged penalty among the candidates; a random positive-change boundary with $\Delta_i>0$; and a random boundary.
The random boundary measures background variation.
The random positive-change boundary asks whether selecting a rise alone explains the effect.
The high-magnitude boundary asks whether entering a block with a high penalty is sufficient to account for the observations.
Appendix~\ref{sec:supp-diagnosis} gives the full protocol.

\begin{figure*}[t]
    \centering
    \includegraphics[width=\textwidth]{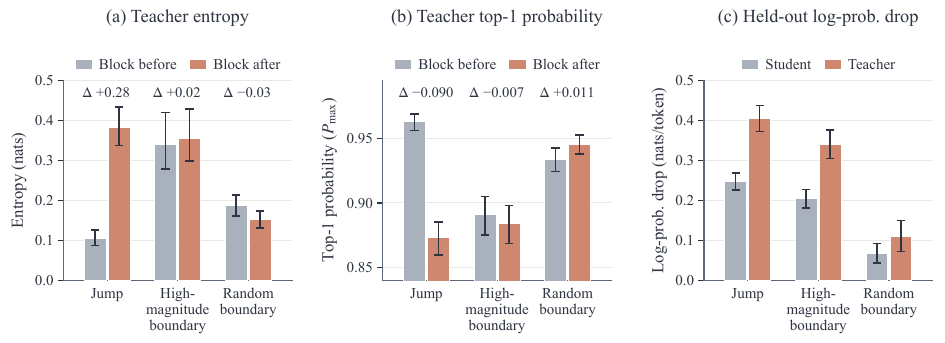}
    \caption{Teacher prediction changes across penalty jumps, abrupt increases in block-averaged penalty detected with the $z=2$ gate, and control boundaries ($B=64$).
    (a) Teacher entropy and (b) teacher top-1 probability, averaged over all tokens in the block before and the block after each boundary, on 153 independent matched groups.
    Their changes indicate decreasing teacher confidence at jumps, accompanying the penalty increase in the observed penalty--confidence tension.
    Appendix~\ref{sec:supp-teacher} also reports all 353 groups.
    $\Delta$ is after minus before, and the vertical axis in (b) is truncated.
    (c) Drop in the student's and the teacher's log-probability of the sampled tokens (before minus after), measured on held-out tokens from 324 prompts; (a,b) use all block tokens.
    Error bars are 95\% bootstrap intervals.}
    \label{fig:boundary}
    \vspace{-9pt} \end{figure*}

\subsection{The Penalty Jumps}
\label{sec:local-rises}
Jumps are selected for a large $\Delta_i$, so measuring the rise on the same tokens would also count the noise that made the boundary pass the gate.
We therefore use a held-out measurement: jumps are detected on the odd-position tokens of each block and the profile is measured on the even-position tokens, and vice versa.
The held-out profiles in Figure~\ref{fig:heldout} show an abrupt penalty increase at jumps, larger than at any of the three controls.
The rise is concentrated in the first block after a jump.
By the second block, the held-out block-averaged penalty is back at the level of a position-matched random boundary.
The analysis in Section~\ref{sec:analysis} shows that the rise persists after token-level clipping.
Appendix~\ref{sec:supp-diagnosis} reports the same pattern in two further training runs.

The post-jump block carries about twice its token share of the penalty: 6.5\% of tokens hold 13.5\% of the penalty mass.
The blocks selected by magnitude-cap are even more concentrated, so concentration alone does not single out jumps.

\subsection{Penalty--Confidence Tension}
\label{sec:teacher}
We measure teacher prediction changes using entropy and top-1 probability separately.
For the teacher distribution $p_t(v)=\pi_T(v\mid c_t)$ over vocabulary $\mathcal V$, these are $H_t=-\sum_{v\in\mathcal V}p_t(v)\log p_t(v)$ and $P_{\max,t}=\max_{v\in\mathcal V}p_t(v)$.
We run full-vocabulary teacher forward passes over the analysis rollouts and report their block means as $H$ and $P_{\max}$.
Panels a and b of Figure~\ref{fig:boundary} show that across the detected jumps, $H$ rises and $P_{\max}$ falls, indicating a decrease in teacher confidence.
Together with the increase in block-averaged penalty, this observed pattern constitutes penalty--confidence tension.
Compared with high-magnitude boundaries, jumps show an additional entropy increase of 0.261 nats and an additional top-1 probability decrease of 0.083.
These paired contrasts compare the post-minus-pre changes within matched groups.
They quantify how teacher predictions change across the boundary, rather than how confident the teacher is in the following block alone.
Local penalty changes therefore reveal differences in teacher predictions beyond entering a block with a high penalty.

Figure~\ref{fig:boundary}c shows that, on the held-out tokens, the log-probabilities that both the student and the teacher assign to the sampled tokens fall across a jump.
For both models, the drop is larger than across either control.
This pattern cannot be explained by an increase in the student's sampled-token probability alone.

\paragraph{Implications for the method.}
Penalty--confidence tension motivates testing whether supervision in the post-jump block should be reduced.
We use the preceding block in the same response as a local reference, without assuming that its penalty is correct or optimal.
The rule in Section~\ref{sec:method} detects penalty jumps rather than teacher uncertainty and does not use teacher confidence as an input.
Training experiments in Section~\ref{sec:ablation} evaluate the intervention: AP-OPD improves performance, whereas magnitude-cap removes a similar amount of penalty without a gain.
This comparison does not establish a causal role for the confidence decrease.

\section{Methodology}
\label{sec:method}
Adaptive Pullback OPD keeps the vanilla advantage $A_t$ and the objective in Eq.~\eqref{eq:objective}, and changes only the token weights $w_t$.
For each response, it computes the block profile in Eq.~\eqref{eq:block-profile} and detects the jumps $\mathcal E$ using Eq.~\eqref{eq:jumps}.
It then uniformly scales supervision in each post-jump block using the preceding block as a local reference.
Algorithm~\ref{alg:ap-opd} gives the complete procedure.

\subsection{Pullback Rule}
\label{sec:pullback}
At a jump $i\in\mathcal E$, AP-OPD uses the preceding block's original block-averaged penalty $D_i$ as a local reference.
It uniformly scales all signed supervision coefficients in the post-jump block by one factor:
\begin{equation}
\renewcommand{\arraystretch}{1.2}
w_t=\left\{
\begin{array}{@{}l@{\;}l@{}}
D_i/D_{i+1}, & t\in\mathcal B_{i+1}\ \text{for some}\ i\in\mathcal E,\\
1, & \text{otherwise}.
\end{array}
\right.
\label{eq:pullback}
\end{equation}
The weighted block-averaged penalty therefore matches the reference:
\begin{equation}
\frac{1}{n_{i+1}}\sum_{t\in\mathcal B_{i+1}}w_t d_t
=\frac{D_i}{D_{i+1}}D_{i+1}=D_i.
\label{eq:pullback-reference}
\end{equation}
A positive weight preserves the signs and relative proportions of the coefficients; a zero weight sets them all to zero.
The scaling also reduces rewarding coefficients when $r_t<0$.
In contrast, setting every $r_t$ to $D_i$ erases token differences and turns negative coefficients into penalties when $D_i>0$; subtracting $\Delta_i$ turns tokens with $0<r_t<\Delta_i$ from penalized to rewarded.

The rule triggers on detected local increases in block-averaged penalty.
Detection and weights both use the observed, unweighted profile.
A block that stays high after a jump therefore keeps weight one.
If the next boundary is also a jump, the post-jump block serves as its pre-jump block, again with its original, unweighted block-averaged penalty.
Each block is weighted at most once.

\subsection{Properties}
\label{sec:properties}
\paragraph{Bounded weights.}
Since $0\le D_i<D_{i+1}$ for every jump, post-jump weights lie in $[0,1)$ and all other weights are one: AP-OPD never upweights a token or flips the sign of a coefficient.

\paragraph{Strength set by the data.}
The penalty mass removed from a post-jump block is $\sum_{t\in\mathcal B_{i+1}}(1-w_t)d_t=n_{i+1}\Delta_i$.
Given the detector and its parameters, the local reference sets the scaling factor without an additional strength constant.
For the default detector, $B=64$ and $z=2$ are fixed before training.
The weights have no additional normalization, clipping, or floor.
Removed penalty mass is a coefficient statistic, not a gradient or parameter-update magnitude.

\paragraph{Selective intervention.}
As reported in Section~\ref{sec:analysis}, about a quarter of training responses contain no jump and are trained exactly as in vanilla OPD.
As quantified in Appendix~\ref{sec:supp-diagnosis}, applying the same ratio at every boundary with $\Delta_i>0$ would remove 36.1\% of the penalty mass instead of 11.6\% under the default detector on the analysis rollouts.

\paragraph{Ranking detector.}
As an alternative to the $z$-gate, the top-$\alpha$ detector ranks the boundaries of each response by $\Delta_i$.
It selects the
\begin{equation}
K=\min\bigl(\lceil\alpha C\rceil,\ |\{i:\Delta_i>0\}|\bigr)
\label{eq:topk}
\end{equation}
largest positive changes, where $C$ is the number of block boundaries and $\alpha=5\%$, and applies the same pullback in Eq.~\eqref{eq:pullback}.

\subsection{Training Objective and Complexity}
\label{sec:objective}
Substituting Eq.~\eqref{eq:pullback} into Eq.~\eqref{eq:objective} gives the AP-OPD objective.
The weights are stop-gradient quantities computed in one $O(T)$ pass over log-probabilities that OPD already produces; they need no additional rollout, model forward pass, or full-vocabulary statistic, and inference is unchanged.
The weights are not renormalized.

\begin{algorithm}[t]
\caption{AP-OPD token weights with the default $z$-gate}
\label{alg:ap-opd}
\begin{algorithmic}[1]
\REQUIRE $r_{1:T}$, block size $B$, threshold $z$
\ENSURE token weights $w_{1:T}$
\STATE $d_t\gets\operatorname{sg}([r_t]_+)$, $w_t\gets1$ for all $t$ \COMMENT{Eq.~\eqref{eq:positive-divergence}}
\STATE Split the tokens into blocks $\mathcal B_{1:M}$
\STATE Compute $D_i$ for each block and $s_i^2$ for each block with $n_i\ge2$ \COMMENT{Eq.~\eqref{eq:block-profile}}
\FOR{$i=1$ \TO $M-1$ with $n_i,n_{i+1}\ge2$}
    \STATE Compute $\Delta_i=D_{i+1}-D_i$ and $\hat\sigma_i$ \COMMENT{Eq.~\eqref{eq:change}}
    \IF{$\Delta_i>z\,\hat\sigma_i$}
        \STATE $w_t\gets D_i/D_{i+1}$ for $t\in\mathcal B_{i+1}$ \COMMENT{Eq.~\eqref{eq:pullback}}
    \ENDIF
\ENDFOR
\RETURN $w_{1:T}$
\end{algorithmic}
\end{algorithm}

\section{Experiments}
\label{sec:experiments}

\begin{table*}[t]
    \centering
    \small
    \setlength{\tabcolsep}{5pt}
    \begin{tabular}{lccccccc}
        \toprule
        Method & MATH500 & AMC23 & Minerva & Olympiad & AIME24 & AIME25 & Macro \\
        \midrule
        Teacher (Qwen3-8B-RL-Math) & 86.80 & 74.22 & 35.75 & 50.19 & 32.08 & 21.67 & 50.12 \\
        Student (Qwen3-1.7B-Base) & 42.15 & 24.06 & 11.67 & 16.63 & 3.54 & 1.67 & 16.62 \\
        \midrule
        Vanilla OPD & 72.90 & 43.75 & 20.96 & \underline{33.63} & \underline{12.40} & 7.29 & 31.82 \\
        IW-OPD & 71.60 & 43.28 & \textbf{22.15} & 32.31 & 11.35 & 7.19 & 31.31 \\
        FiRe-OPD & 72.80 & \underline{45.39} & 21.74 & 32.83 & 11.88 & \underline{8.44} & 32.18 \\
        \midrule
        AP-OPD (top-$\alpha$) & \underline{72.97} & 44.77 & \underline{22.06} & \textbf{34.26} & \textbf{13.02} & \textbf{8.65} & \textbf{32.62} \\
        AP-OPD ($z$-gate) & \textbf{73.08} & \textbf{46.72} & 21.42 & 32.94 & 12.19 & 8.02 & \underline{32.39} \\
        \bottomrule
    \end{tabular}
    \caption{Main results at step 200, reported as Avg@$k$ in percent under a shared evaluation protocol, averaged over two training runs.
    The teacher and the student before distillation are reference evaluations and are not ranked.
    Best in bold, second underlined; Macro weights the six benchmarks equally.}
    \label{tab:main}
    \vspace{-6pt} \end{table*}

\subsection{Experimental Setup}
\label{sec:setup}
\paragraph{Training.}
For mathematical reasoning, the student is Qwen3-1.7B-Base \citep{yang2025qwen3}.
The frozen teacher is Qwen3-8B-RL-Math, which is Qwen3-8B-Base trained on mathematics with GSPO \citep{zheng2025gspo}.
Training prompts come from DAPO-Math-17K \citep{yu2025dapo}, and each method trains for 200 steps.
For code generation, Qwen3-4B-Instruct-2507 is distilled into Qwen3-1.7B on the code subset of Eurus-2-RL-Data \citep{cui2025prime}, and each method trains for 100 steps.
Within each domain, every method starts from the same student checkpoint.
In both domains, our runs use 128 prompts per step, one rollout per prompt of at most 8{,}192 tokens, and AdamW with a constant learning rate of $10^{-6}$.
AP-OPD uses $B=64$ and $z=2$ in both domains, fixed before training.
Details are shown in Appendix~\ref{sec:supp-setup}.

\paragraph{Baselines.}
We compare with vanilla token-level OPD \citep{fu2026revisiting} and two supervision reweighting methods.
IW-OPD uses accumulated absolute log-ratios, normalized within each response, to assign decreasing prefix weights \citep{xie2026iwopd}.
FiRe-OPD filters trajectories and reweights tokens by teacher and student confidence \citep{li2026fireopd}.
Both use their authors' code and default optimization settings, and share the student, teacher, training prompts, number of steps, and evaluation with our runs.
On code generation, we compare with vanilla OPD and IW-OPD.

\paragraph{Evaluation.}
For mathematical reasoning, we report Avg@$k$, the accuracy averaged over $k$ sampled responses per problem, on MATH500 \citep{hendrycks2021math,lightman2024verify}, AMC23, Minerva \citep{lewkowycz2022minerva}, OlympiadBench \citep{he2024olympiadbench}, AIME24, and AIME25.
Answers are graded with the rule-based math checker of verl \citep{sheng2025hybridflow}.
For code generation, we report Avg@4 on HumanEval+, MBPP+ \citep{liu2023evalplus}, and LiveCodeBench v6 \citep{jain2024livecodebench}, running each program against the benchmark's tests.
All models share one decoding setup.
Mathematical results average two training runs that differ in data order.
Appendix~\ref{sec:supp-setup} gives the full protocol, and Appendix~\ref{sec:supp-results} reports each run.

\subsection{Main Results}
\label{sec:main-results}
\paragraph{Mathematical reasoning.}
Table~\ref{tab:main} reports AP-OPD with each of its two jump detectors, the $z$-gate of Section~\ref{sec:block-profile} and the top-$\alpha$ ranking of Eq.~\eqref{eq:topk}.
Both detectors improve the macro average over vanilla OPD, by 0.80 points for top-$\alpha$ and 0.57 for the $z$-gate.
The improvement holds in each training run (Appendix~\ref{sec:supp-results}).
In the two-run mean, top-$\alpha$ improves on every benchmark.
Both detectors also obtain higher macro averages than IW-OPD and FiRe-OPD in this mean.
FiRe-OPD, the stronger baseline, improves on vanilla OPD by less than either detector.
Computing its token weights requires the full-vocabulary entropies of the teacher and the student.
AP-OPD uses only the sampled-token log-ratios that OPD already computes.

\paragraph{Choice of detector.}
The selected boundary sets of the two detectors have a Jaccard overlap of about one third on the analysis rollouts.
Since both improve over vanilla OPD, the gain is not tied to one detector.
We keep the $z$-gate as the default as it needs no selection ratio and leaves responses without a clear rise untouched.

\begin{table}[t]
    \centering
    \small
    \begin{tabular*}{\columnwidth}{@{\extracolsep{\fill}}lcccc@{}}
        \toprule
        Method & HE+ & MBPP+ & LCB & Macro \\
        \midrule
        Teacher & 83.69 & 72.95 & 32.29 & 62.97 \\
        Student & 61.59 & 56.81 & 14.57 & 44.32 \\
        \midrule
        Vanilla OPD & 70.43 & 63.89 & 23.57 & 52.63 \\
        IW-OPD & 70.88 & \textbf{65.28} & 25.00 & 53.72 \\
        AP-OPD & \textbf{73.17} & 65.15 & \textbf{25.71} & \textbf{54.68} \\
        \bottomrule
    \end{tabular*}
    \caption{Code generation uses Avg@4 (\%).
    Qwen3-4B-Instruct-2507 (Teacher) and undistilled Qwen3-1.7B (Student) are unranked references.
    HE+ is HumanEval+; LCB is LiveCodeBench v6.
    Macro is the unweighted mean; best trained scores are bold.}
    \label{tab:code}
    \vspace{-6pt} \end{table}

\paragraph{Code generation.}
Code generation tests AP-OPD with a different student, teacher, and training data.
The student is already post-trained.
The teacher is a general instruction model.
With the default $z$-gate, AP-OPD obtains the highest macro average, 2.05 points above vanilla OPD and 0.96 above IW-OPD (Table~\ref{tab:code}).
It improves over vanilla OPD on all three benchmarks.
It leads on HumanEval+ and on LiveCodeBench, the hardest benchmark, and is close to IW-OPD on MBPP+.
The gain over vanilla OPD is larger than on mathematical reasoning.
Across both domains, down-weighting supervision after detected jumps improves over vanilla OPD.
Section~\ref{sec:ablation} compares AP-OPD with magnitude-cap to test whether removing a similar amount of penalty is sufficient to obtain the observed gains.

\suppressfloats[t]
\begin{table}[t]
    \centering
    \small
    \setlength{\tabcolsep}{3pt}
    \begin{tabular}{lccccc}
        \toprule
        Variant & \multicolumn{3}{c}{Offline} & \multicolumn{2}{c}{Trained} \\
        \cmidrule(lr){2-4}\cmidrule(lr){5-6}
         & Cov. & Unsel. & $q$ & Macro & $\Delta$ \\
        \midrule
        \multicolumn{6}{l}{(a) Comparison with magnitude-cap} \\
        Vanilla OPD & -- & -- & -- & 31.55 & -- \\
        Magnitude-cap & 7.17 & 0.00 & 13.75 & 31.40 & $-0.15$ \\
        AP-OPD (top-$\alpha$) & 7.19 & 0.00 & 14.63 & \textbf{32.50} & $+0.95$ \\
        AP-OPD ($z$-gate) & 6.49 & 28.71 & 11.60 & \textbf{32.50} & $+0.95$ \\
        \midrule
        \multicolumn{6}{l}{(b) Block size, $z$-gate} \\
        $B=32$ & 4.23 & 24.22 & 9.51 & 32.06 & $+0.50$ \\
        $B=64$ & 6.49 & 28.71 & 11.60 & \textbf{32.50} & $+0.95$ \\
        $B=128$ & 8.60 & 36.72 & 11.78 & 31.96 & $+0.40$ \\
        \bottomrule
    \end{tabular}
    \caption{Ablations, same number of training steps and evaluation protocol as Table~\ref{tab:main}, first run only.
    Macro: macro Avg@$k$; $\Delta$: difference from vanilla OPD.
    Offline statistics on the analysis rollouts are reported in percent: Cov., response-mean share of tokens in selected blocks; Unsel., share of responses with no selected boundary; $q$, response-mean fraction of penalty mass removed.}
    \label{tab:ablation}
    \vspace{-6pt} \end{table}

\subsection{Ablation Studies}
\label{sec:ablation}
\paragraph{Comparison with magnitude-cap.}
Magnitude-cap selects the $\lceil\alpha C\rceil$ post-boundary blocks with the largest $D_{i+1}$ and assigns weight $\min(1,\bar D/D_{i+1})$, where $\bar D$ is the response's mean penalty.
We measure the removed fraction as $q=\sum_t(1-w_t)d_t/\sum_t d_t$, averaged equally over responses with nonzero penalty mass.
Magnitude-cap removes 13.75\%, between the two AP-OPD detectors, but brings no gain and trails both by about one point (Table~\ref{tab:ablation}a; paired intervals in Appendix~\ref{sec:supp-results}).
An earlier experiment also gains from a fixed post-jump weight of 0.3, whereas pre-jump upweighting brings no gain (Appendix~\ref{sec:supp-negative}).
The tested gains thus occur with two detectors and two post-jump weights; a similar removal amount alone is insufficient.
These rule comparisons do not isolate selection, reference, and scaling depth.

\paragraph{Block size.}
The block-size ablation in Table~\ref{tab:ablation}b shows improvements over vanilla OPD and magnitude-cap for all three block sizes.
Larger blocks select fewer jumps but weight longer spans, so the removed mass $q$ stays within 9.5--11.8\%.
The default $B=64$ gives the largest gain.
Shorter blocks may make block means noisier, while longer blocks may average over local increases.

\subsection{Analysis}
\label{sec:analysis}
\paragraph{Relation to token-level clipping.}
The rise is not carried by a few extreme tokens, so clipping each token's log-ratio, as REOPOLD does \citep{ko2026reopold}, leaves it largely in place.
At a clipping bound that removes as much penalty mass as AP-OPD on the analysis set, jumps are still detected and about 90\% of the held-out rise remains; REOPOLD's default bound removes more than twice as much and still keeps about 70\%.
The two rules also remove different supervision.
Appendix~\ref{sec:supp-reopold} reports that only 14.5\% of AP-OPD's removal lies above the bound, close to the 12.5\% expected if its removal were unrelated to token size.

\paragraph{Training statistics.}
During training, AP-OPD marks 6.8\% of boundaries as jumps and leaves 24.2\% of responses unweighted.
Gradient clipping fires less often than in vanilla OPD.
The training statistics in Appendix~\ref{sec:supp-results} show that response length and student entropy stay close to vanilla OPD.
The mean token weight is 0.945.
Magnitude-cap has a similar mean weight, 0.961, but brings no gain.
The removal amount alone therefore does not explain the performance difference.

\section{Related Work}
\label{sec:related}

\paragraph{On-policy distillation.}
Unlike distillation from fixed or teacher-generated data \citep{hinton2015distilling,kim2016sequence,hsieh2023distilling,wang2021selective,xie2026adakd}, OPD scores the student's own rollouts with the teacher \citep{agarwal2024gkd}.
GKD supports on-policy learning with alternative loss functions \citep{agarwal2024gkd}.
MiniLLM minimizes reverse KL with policy gradients to prevent overestimating low-probability regions of the teacher distribution \citep{gu2024minillm}.
DistiLLM uses an adaptive off-policy approach to reuse student-generated outputs \citep{ko2024distillm}.
Recent work extends OPD to reasoning post-training \citep{zhu2026hybrid,zhang2026prefixopd,ko2026reopold,li2026rethinkingopd} and studies its supervision and optimization \citep{song2026opdsurvey,jang2026veto,jin2026entropyopd,yang2026exopd,zhang2026truncatedopd,wang2026tcod}.
RevisitingOPD documents failure modes of sampled-token comparison, including unreliable guidance on student-drifted prefixes \citep{fu2026revisiting}.

\paragraph{Reweighting, pruning, and down-weighting in OPD.}
TIP selects tokens by student uncertainty and teacher--student divergence \citep{xu2026tip}.
FiRe-OPD filters trajectories and softly reweights tokens by teacher and student confidence \citep{li2026fireopd}.
IW-OPD uses accumulated absolute log-ratios, normalized within each response, to assign decreasing prefix weights \citep{xie2026iwopd}.
Prune-OPD decays the supervision of a suffix once teacher--student support mismatch accumulates \citep{yang2026pruneopd}.
REOPOLD clips token rewards at a fixed bound \citep{ko2026reopold}.
Rock Tokens studies tokens whose distillation loss persists over training \citep{jiang2026rocktokens}, Blockwise Gating weights 64-token blocks by student policy drift \citep{zheng2026blockwise}, and entropy-aware OPD adds a forward-KL term where teacher entropy is high \citep{jin2026entropyopd}.
TrOPD gates the penalty by the teacher--student probability ratio \citep{xing2026tropd}.
AP-OPD instead uses changes between consecutive blocks in the same response and scales post-jump supervision using the preceding block as a local reference.
Unlike confidence-based weighting, its rule uses penalty changes without measuring teacher confidence.

\paragraph{Temporal information.}
Return-to-go OPD propagates later log-ratios to earlier positions, at the cost of higher gradient variance \citep{gu2024minillm,fu2026revisiting}.
TOPD and AC-OPD obtain near-future evidence by generating continuations from selected prefixes \citep{jiang2026trajectoryopd,zhou2026acopd}.
LGR evaluates the student's top-K candidate tokens by the one-step-ahead teacher confidence they induce and adds a group-normalized reward \citep{liu2026lgr}.
AP-OPD uses only changes realized along the completed rollout and needs no additional generation or forward pass.

\paragraph{Choice of baselines.}
We compare with IW-OPD and FiRe-OPD as supervision reweighting baselines on student-generated rollouts; FiRe-OPD also filters trajectories.
Other methods introduce changes to objectives, rollout generation, or token selection.
TrOPD and entropy-aware OPD add a forward-KL term, and TrOPD also uses teacher-generated prefixes.
LGR, TOPD, and AC-OPD add teacher passes or continuations.
Prune-OPD truncates rollouts, and TIP keeps a subset of tokens to reduce training memory.
Rock Tokens selects token types by their loss over training.
Blockwise Gating gates on the drift between the rollout and the current student, which is zero when each batch is used for one update, as in our setting.
REOPOLD also adds stage-dependent token masks, so we compare only its token-level clipping, directly on our rollouts (\S\ref{sec:analysis}).
Magnitude-cap selects blocks by their block-averaged penalties and removes a similar amount of penalty mass, trained with the same configuration as AP-OPD (\S\ref{sec:ablation}).

\section{Conclusion}
We studied local penalty changes in on-policy distillation, where the current penalty value does not distinguish sustained high penalties from abrupt increases along a response.
Comparing consecutive token blocks reveals penalty jumps that persist on held-out tokens.
Across these jumps, penalties rise while teacher confidence decreases, whereas boundaries into high-penalty blocks do not exhibit comparable confidence changes.
This penalty--confidence tension shows that local changes provide information beyond the current penalty value.
Motivated by this observation, we proposed AP-OPD, which uniformly scales supervision coefficients in each post-jump block using the preceding block as a local reference.
The weighted average penalty matches the preceding block's original mean, setting the scaling factor without an additional strength constant.
AP-OPD improves performance on mathematical reasoning and code generation.

\clearpage
\section*{Limitations}
We evaluate AP-OPD on mathematical reasoning and code generation, and have not yet applied it to other settings such as agentic or multi-turn tasks.
Our diagnostic evidence for penalty--confidence tension is observational and does not establish that decreased teacher confidence causes the gains from pullback.
AP-OPD also leaves room for improvement in how it forms blocks.
It uses a fixed block length, and adapting the block size to each response or aligning blocks with reasoning steps is a natural extension.
Finally, AP-OPD responds only to abrupt rises, so it does not address reward-hacking-like behavior in OPD such as repetitive looping, where the penalty fades rather than jumps.

\begingroup
\let\emph\textnormal
\bibliography{references}
\endgroup

\clearpage
\appendix
\raggedbottom
\makeatletter
\setlength{\@dblfptop}{0pt}
\setlength{\@dblfpsep}{16pt}
\setlength{\@dblfpbot}{0pt plus 1fil}
\makeatother

\section{Experimental Details}
\label{sec:supp-setup}

\subsection{Student Training}
\label{sec:supp-training}
Table~\ref{tab:supp-training} lists the mathematical-reasoning configuration shared by vanilla OPD, both AP-OPD detectors, magnitude-cap, and the block-size runs; the code experiment is described in Appendix~\ref{sec:supp-code}.
\begin{table}[H]
    \centering
    \small
    \setlength{\tabcolsep}{3pt}
    \begin{tabularx}{\columnwidth}{@{}L{0.33\columnwidth}Y@{}}
        \toprule
        Setting & Value \\
        \midrule
        Student initialization & Qwen3-1.7B-Base \\
        Frozen teacher & Qwen3-8B-RL-Math; see Table~\ref{tab:supp-teacher-training} \\
        Training prompts & DAPO-Math-17K; 17{,}398 prompts \\
        Prompt / response limit & 2{,}048 / 8{,}192 tokens \\
        Steps & 200, from the base checkpoint \\
        Batch & 128 prompts, one rollout each; one update per step \\
        Rollout sampling & temperature 1.0, top-$p$ 1.0, no top-$k$; thinking disabled \\
        Optimizer & AdamW, $\beta=(0.9,0.999)$, weight decay 0.01 \\
        Learning rate & $10^{-6}$, constant, no warmup \\
        Gradient clipping & global norm 1.0 \\
        Loss & token mean; no task reward, KL loss, or entropy bonus \\
        Data order & not shuffled; run 1: data seed 42; run 2: data seed 43 \\
        AP-OPD & $B=64$, $z=2$; top-$\alpha$ uses $\alpha=5\%$ \\
        \bottomrule
    \end{tabularx}
    \caption{Training configuration shared by our mathematical-reasoning runs.}
    \label{tab:supp-training}
\end{table}

\subsection{Teacher}
\label{sec:supp-teacher-training}
The mathematical-reasoning teacher, Qwen3-8B-RL-Math, is Qwen3-8B-Base trained with GSPO \citep{zheng2025gspo} on DAPO-Math-17K.
Table~\ref{tab:supp-teacher-training} lists its configuration.
The teacher is frozen during distillation.

\begin{table}[H]
    \centering
    \small
    \setlength{\tabcolsep}{3pt}
    \begin{tabularx}{\columnwidth}{@{}L{0.33\columnwidth}Y@{}}
        \toprule
        Setting & Value \\
        \midrule
        Initialization & Qwen3-8B-Base \\
        Training prompts & DAPO-Math-17K \\
        Policy optimization & GSPO \\
        Steps & 200 \\
        Prompt / response limit & 2{,}048 / 4{,}096 tokens \\
        Batch & 512 prompts, 16 responses each; mini-batch of 32 prompts \\
        Clipping range & $3\times10^{-4}$ lower, $4\times10^{-4}$ upper \\
        Learning rate & $10^{-6}$, 10 warmup steps \\
        Weight decay & 0.1 \\
        Reward & $+1$ for a correct answer, $-1$ otherwise; no overlong shaping \\
        \bottomrule
    \end{tabularx}
    \caption{Training configuration of the frozen mathematical-reasoning teacher.}
    \label{tab:supp-teacher-training}
\end{table}

\subsection{Evaluation}
\label{sec:supp-evaluation}
All models use the evaluation configuration in Table~\ref{tab:supp-eval}, with 7{,}388 responses per model.
Prompts are capped at 2{,}048 tokens and responses at 16{,}384; sampling uses temperature 0.7, top-$p$ 0.8, top-$k$ 20, and seed 42, with thinking mode disabled.
For a benchmark with $N$ problems and correctness $c_{q,j}\in\{0,1\}$ of sample $j$ on problem $q$,
\begin{equation}
\operatorname{Avg@}k=\frac{1}{Nk}\sum_{q=1}^{N}\sum_{j=1}^{k}c_{q,j}.
\end{equation}
Answers are graded with the rule-based math checker of verl.

\paragraph{Paired bootstrap.}
For a contrast between two models, we resample the problems of each benchmark with replacement 10{,}000 times using seed 42, keep the existing $k$ scores of each resampled problem for both models, and recompute the per-benchmark and macro differences.
We report percentile intervals.
The intervals use the run-1 checkpoints and cover evaluation randomness, not training variance.

\begin{table}[H]
    \centering
    \small
    \begin{tabular}{lrrl}
        \toprule
        Benchmark & Problems & $k$ & Metric \\
        \midrule
        MATH500 & 500 & 4 & Avg@4 \\
        AMC23 & 40 & 16 & Avg@16 \\
        Minerva & 272 & 4 & Avg@4 \\
        OlympiadBench & 675 & 4 & Avg@4 \\
        AIME24 & 30 & 16 & Avg@16 \\
        AIME25 & 30 & 16 & Avg@16 \\
        \bottomrule
    \end{tabular}
    \caption{Evaluation benchmarks and samples per problem.}
    \label{tab:supp-eval}
\end{table}

\subsection{Code Generation}
\label{sec:supp-code}
The code experiment distills Qwen3-4B-Instruct-2507 into Qwen3-1.7B, with the configuration in Table~\ref{tab:supp-code-training}.
Training prompts are drawn from the 25{,}276 training problems in the code subset of Eurus-2-RL-Data \citep{cui2025prime}, competitive-programming problems with test cases.
All methods train for 100 steps from the same student checkpoint, and we evaluate the step-100 checkpoints.
We evaluate on 164 HumanEval+ problems, 378 MBPP+ problems, and 175 LiveCodeBench v6 problems, with four samples per problem, responses capped at 16{,}384 tokens, temperature 0.7, top-$p$ 0.8, top-$k$ 20, and thinking disabled.
Each program is executed in a sandbox against the benchmark's tests, and Avg@4 is the share of passing samples.

\begin{table}[H]
    \centering
    \small
    \setlength{\tabcolsep}{3pt}
    \begin{tabularx}{\columnwidth}{@{}L{0.33\columnwidth}Y@{}}
        \toprule
        Setting & Value \\
        \midrule
        Student initialization & Qwen3-1.7B \\
        Frozen teacher & Qwen3-4B-Instruct-2507 \\
        Training prompts & Eurus-2-RL-Data code subset; 12{,}800 prompts \\
        Prompt / response limit & 2{,}048 / 8{,}192 tokens \\
        Steps & 100, one epoch \\
        Batch & 128 prompts, one rollout each; one update per step \\
        Rollout sampling & temperature 1.0, top-$p$ 1.0, no top-$k$; thinking disabled \\
        Optimizer & AdamW, $\beta=(0.9,0.999)$, weight decay 0.01 \\
        Learning rate & $10^{-6}$, constant, no warmup \\
        Gradient clipping & global norm 1.0 \\
        Task reward & none \\
        AP-OPD & $z$-gate, $B=64$, $z=2$ \\
        \bottomrule
    \end{tabularx}
    \caption{Training configuration of our code-generation runs.}
    \label{tab:supp-code-training}
\end{table}

\FloatBarrier
\section{Diagnostic Analysis Details}
\label{sec:supp-diagnosis}

\subsection{Rollouts}
The analysis set contains 512 rollouts, one per distinct prompt, saved by the vanilla OPD run at steps 70, 80, 90, and 100, with 128 rollouts per step.
For replication we use two further training runs with rollouts saved every 10 steps over 300 steps, with 3{,}840 rollouts per run: a vanilla OPD run and a run trained with the paired rule of Appendix~\ref{sec:supp-negative}.
All statistics are computed offline and do not affect training.

\subsection{Detector Statistics}
Table~\ref{tab:supp-detector} compares jump detectors and other selection rules on the analysis set.
The two detectors select overlapping but different boundaries, with a Jaccard index of 0.33.

\begin{table*}[bp]
    \centering
    \small
    \setlength{\tabcolsep}{6pt}
    \begin{tabular}{lcccc}
        \toprule
        Rule & Selected/resp. & Unselected (\%) & Token cov. (\%) & $q$ (\%) \\
        \midrule
        Top-$\alpha$ ($5\%$) & 1.85 & 0.00 & 7.19 & 14.63 \\
        $z=1.5$ & 3.93 & 8.98 & 13.37 & 20.04 \\
        $z=2$ (AP-OPD) & 1.91 & 28.71 & 6.49 & 11.60 \\
        $z=3$ & 0.36 & 75.78 & 1.10 & 2.56 \\
        All $\Delta_i>0$ & -- & -- & -- & 36.11 \\
        Magnitude-cap & 1.85 & 0.00 & 7.17 & 13.75 \\
        \bottomrule
    \end{tabular}
    \caption{Selection rules on the analysis rollouts at $B=64$: selected boundaries per response, share of responses without a selection, share of tokens in selected blocks, and removed penalty fraction $q$.
    Token coverage and $q$ are averaged equally over responses; $q$ uses responses with nonzero penalty mass.
    Among the $z$-gate thresholds, only $z=2$ is used for training; the other $z$ values are reported for reference.
    Magnitude-cap selections are not penalty jumps.}
    \label{tab:supp-detector}
\end{table*}

\subsection{Split-Half Protocol}
\label{sec:supp-splithalf}
Profiles use the window $[-2B,2B)$ around each boundary, with offset 0 at the first token of the following block, and never cross a mask break or the response boundary.
Controls come from the same response and the same decile of relative position, defined as the boundary offset divided by valid length.
Within each control type, boundaries are drawn without replacement; the high-magnitude boundary is the candidate with the highest $D_{i+1}$.
For split-half measurement, detection uses only the odd-indexed tokens of each 64-token block, including the token counts in the gate, and controls are also selected on odd tokens; all quantities are then measured on even tokens.
The reverse direction swaps the roles, and the two directions are averaged with shared bootstrap weights.
Estimates are averaged within each response, then over prompts; intervals come from 10{,}000 prompt-clustered resamples using seed 42.
The held-out rise $U_{a:b}$ is the measured mean of $d_t$ over offsets $[a,b)$ minus the measured mean of the block preceding the selected boundary.
For a jump, this reference is the pre-jump block; each control uses its own preceding block.
On the analysis set, 910 of the 978 $z=2$ jumps have a complete window, covering 346 responses; with odd and even detection there are 478 and 453 complete jumps from 247 and 230 responses, and 324 prompts in the union.
Each detection half of a block has only 32 tokens, and adjacent tokens remain correlated, so split-half measurement reduces but does not remove selection effects.

\begin{table*}[!tbp]
    \centering
    \small
    \setlength{\tabcolsep}{6pt}
    \begin{tabular}{lccc}
        \toprule
        Detect & $-$ High-mag.\ bound. & $-$ Rand.\ pos. & $-$ Rand.\ bound. \\
        \midrule
        odd & 0.019 {\scriptsize[0.006, 0.031]} & 0.041 {\scriptsize[0.025, 0.058]} & 0.135 {\scriptsize[0.105, 0.168]} \\
        even & 0.036 {\scriptsize[0.016, 0.057]} & 0.050 {\scriptsize[0.034, 0.066]} & 0.145 {\scriptsize[0.116, 0.177]} \\
        \bottomrule
    \end{tabular}
    \caption{Held-out rise $U_{0:64}$ of $z=2$ jumps minus that of each matched control, for each detection direction on the analysis set at $B=64$, with 95\% intervals.}
    \label{tab:supp-contrasts}
\end{table*}

\begin{table*}[!tbp]
    \centering
    \small
    \setlength{\tabcolsep}{6pt}
    \begin{tabular}{lccc}
        \toprule
        Rollouts & $U_{0:64}$ & $U_{64:128}$ & Ratio \\
        \midrule
        Analysis set & 0.190 {\scriptsize[0.172, 0.209]} & 0.096 {\scriptsize[0.081, 0.111]} & 50.3\% \\
        Vanilla, 300 steps & 0.188 {\scriptsize[0.183, 0.194]} & 0.101 {\scriptsize[0.096, 0.106]} & 53.6\% \\
        Paired rule, 300 steps & 0.190 {\scriptsize[0.184, 0.196]} & 0.103 {\scriptsize[0.098, 0.108]} & 54.5\% \\
        \bottomrule
    \end{tabular}
    \caption{Held-out rise of the first and second blocks after $z=2$ jumps and their ratio, on the analysis set and on two further training runs.
    Both rises use the pre-jump block as the reference and are measured with the split-half protocol, averaging both detection directions.
    A positive second-block rise relative to this reference is compatible with returning to the level of position-matched random boundaries in Figure~\ref{fig:heldout}.}
    \label{tab:supp-replication}
\end{table*}

\subsection{Teacher Confidence}
\label{sec:supp-teacher}
Of the 512 analysis prompts, 362 contain an eligible $z=2$ jump, and 353 of these yield a complete group with a position-matched high-magnitude boundary and a random boundary.
153 groups share no prompt or control prefix with one another within the subset; the full set comprises 80 connected components.
The teacher is run causally over the original student history, without re-encoding prefixes or truncating early context, and $H$ and $P_{\max}$ are averaged over the 64 tokens of each block.
Intervals resample the 80 components for all groups and resample groups within training steps for the independent subset, using 10{,}000 resamples with seed 42.
Table~\ref{tab:supp-did} reports both sets.
These distributional changes support the observed penalty--confidence tension for the default $z=2$ jump set; they are not held-out measurements or a combined confidence score.

\begin{table*}[!tbp]
    \centering
    \small
    \setlength{\tabcolsep}{6pt}
    \begin{tabular}{llcc}
        \toprule
        Metric & Control & All 353 groups & Independent 153 \\
        \midrule
        $H$ (nats) & random & 0.300 {\scriptsize[0.275, 0.342]} & 0.311 {\scriptsize[0.263, 0.365]} \\
        $H$ (nats) & high-magnitude & 0.172 {\scriptsize[0.122, 0.265]} & 0.261 {\scriptsize[0.212, 0.317]} \\
        $P_{\max}$ & random & $-$0.099 {\scriptsize[$-$0.112, $-$0.091]} & $-$0.102 {\scriptsize[$-$0.116, $-$0.088]} \\
        $P_{\max}$ & high-magnitude & $-$0.061 {\scriptsize[$-$0.085, $-$0.050]} & $-$0.083 {\scriptsize[$-$0.098, $-$0.069]} \\
        \bottomrule
    \end{tabular}
    \caption{DID of the teacher's entropy and top-1 probability, computed as the post-minus-pre change at jumps minus that at control boundaries, with 95\% intervals.}
    \label{tab:supp-did}
\end{table*}

\paragraph{Log-probability decomposition.}
Writing $\Delta S$ and $\Delta T$ for the post-minus-pre changes in the block means of $\ell_t^S$ and $\ell_t^T$, the change in the mean signed log-ratio is $\Delta S-\Delta T$.
This identity does not directly give the change in block-averaged penalty, which averages $[r_t]_+$.
In the held-out setting of Figure~\ref{fig:boundary}c, the teacher's drop exceeds that at high-magnitude boundaries by 0.052 $[0.028, 0.077]$ with odd detection and 0.077 $[0.043, 0.112]$ with even detection; the student's drop also exceeds that of both controls.

\FloatBarrier
\section{Supplementary Results}
\label{sec:supp-results}
Table~\ref{tab:supp-all-scores} lists the per-benchmark scores of all trained models, and Table~\ref{tab:supp-paired} gives per-benchmark paired intervals for the contrasts among our own runs.
The five methods in the main comparison are reported for each run and for their mean.
Magnitude-cap and the additional block sizes were trained in run 1 only.
The paired intervals and training dynamics also refer to run 1.

\begin{table*}[tb]
    \centering
    \small
    \setlength{\tabcolsep}{5pt}
    \begin{tabular}{lccccccc}
        \toprule
        Model & MATH500 & AMC23 & Minerva & Olympiad & AIME24 & AIME25 & Macro \\
        \midrule
        \multicolumn{8}{l}{Mean of the two runs} \\
        Vanilla OPD & 72.90 & 43.75 & 20.96 & \underline{33.63} & \underline{12.40} & 7.29 & 31.82 \\
        IW-OPD & 71.60 & 43.28 & \textbf{22.15} & 32.31 & 11.35 & 7.19 & 31.31 \\
        FiRe-OPD & 72.80 & \underline{45.39} & 21.74 & 32.83 & 11.88 & \underline{8.44} & 32.18 \\
        AP-OPD (top-$\alpha$) & \underline{72.97} & 44.77 & \underline{22.06} & \textbf{34.26} & \textbf{13.02} & \textbf{8.65} & \textbf{32.62} \\
        AP-OPD ($z$-gate, $B=64$) & \textbf{73.08} & \textbf{46.72} & 21.42 & 32.94 & 12.19 & 8.02 & \underline{32.39} \\
        \midrule
        \multicolumn{8}{l}{Run 1} \\
        Vanilla OPD & \textbf{73.50} & 43.75 & 20.22 & 33.30 & 12.08 & 6.46 & 31.55 \\
        IW-OPD & \underline{73.05} & 43.91 & \textbf{21.60} & 31.78 & 11.46 & \textbf{8.75} & 31.76 \\
        FiRe-OPD & 73.00 & 44.22 & \underline{21.42} & 33.00 & 11.46 & 8.12 & \underline{31.87} \\
        AP-OPD (top-$\alpha$) & \underline{73.05} & \underline{45.94} & \underline{21.42} & \textbf{34.41} & \textbf{13.12} & 7.08 & \textbf{32.50} \\
        AP-OPD ($z$-gate, $B=64$) & 72.75 & \textbf{46.56} & 21.32 & \underline{33.33} & \underline{12.71} & \underline{8.33} & \textbf{32.50} \\
        \midrule
        \multicolumn{8}{l}{Run 2} \\
        Vanilla OPD & 72.30 & 43.75 & 21.69 & \underline{33.96} & \underline{12.71} & 8.12 & 32.09 \\
        IW-OPD & 70.15 & 42.66 & \textbf{22.70} & 32.85 & 11.25 & 5.62 & 30.87 \\
        FiRe-OPD & 72.60 & \underline{46.56} & \underline{22.06} & 32.67 & 12.29 & \underline{8.75} & \underline{32.49} \\
        AP-OPD (top-$\alpha$) & \underline{72.90} & 43.59 & \textbf{22.70} & \textbf{34.11} & \textbf{12.92} & \textbf{10.21} & \textbf{32.74} \\
        AP-OPD ($z$-gate, $B=64$) & \textbf{73.40} & \textbf{46.88} & 21.51 & 32.56 & 11.67 & 7.71 & 32.29 \\
        \midrule
        \multicolumn{8}{l}{Additional ablations} \\
        Magnitude-cap & \underline{72.55} & \underline{44.53} & 21.88 & 32.56 & 9.58 & 7.29 & 31.40 \\
        AP-OPD ($z$-gate, $B=32$) & 71.30 & \textbf{45.16} & \underline{22.61} & \underline{32.85} & \textbf{12.29} & \textbf{8.13} & \textbf{32.06} \\
        AP-OPD ($z$-gate, $B=128$) & \textbf{73.15} & 42.81 & \textbf{22.79} & \textbf{33.81} & \underline{11.25} & \underline{7.92} & \underline{31.96} \\
        \bottomrule
    \end{tabular}
    \caption{Per-benchmark Avg@$k$ in percent at step 200 under the same evaluation protocol.
    Mean: arithmetic mean of the two runs, calculated from unrounded scores.
    Macro: equal-weight mean over the six benchmarks; the additional ablations use run 1 only.
    Best in bold, second underlined within each block; ties at the reported precision share the same marking.}
    \label{tab:supp-all-scores}
\end{table*}

\begin{table*}[!t]
    \centering
    \small
    \setlength{\tabcolsep}{4pt}
    \begin{tabular}{lccccccc}
        \toprule
        Contrast & MATH500 & AMC23 & Minerva & Olympiad & AIME24 & AIME25 & Macro \\
        \midrule
        $z$-gate $-$ Vanilla & $-0.75$ & $+2.81$ & $+1.10$ & $+0.04$ & $+0.63$ & $+1.88$ & $+0.95$ \\
        & {\scriptsize[$-$2.40, 0.80]} & {\scriptsize[$-$0.47, 6.25]} & {\scriptsize[$-$1.19, 3.31]} & {\scriptsize[$-$1.44, 1.48]} & {\scriptsize[$-$1.04, 2.29]} & {\scriptsize[$-$0.42, 5.00]} & {\scriptsize[0.02, 1.90]} \\
        Top-$\alpha$ $-$ Vanilla & $-0.45$ & $+2.19$ & $+1.19$ & $+1.11$ & $+1.04$ & $+0.63$ & $+0.95$ \\
        & {\scriptsize[$-$2.05, 1.10]} & {\scriptsize[$-$0.94, 5.47]} & {\scriptsize[$-$0.92, 3.31]} & {\scriptsize[$-$0.33, 2.52]} & {\scriptsize[$-$1.67, 4.17]} & {\scriptsize[$-$1.25, 2.71]} & {\scriptsize[0.01, 1.90]} \\
        Magnitude-cap $-$ Vanilla & $-0.95$ & $+0.78$ & $+1.65$ & $-0.74$ & $-2.50$ & $+0.83$ & $-0.15$ \\
        & {\scriptsize[$-$2.60, 0.65]} & {\scriptsize[$-$1.56, 3.28]} & {\scriptsize[$-$0.46, 3.86]} & {\scriptsize[$-$2.22, 0.70]} & {\scriptsize[$-$5.00, $-$0.21]} & {\scriptsize[$-$1.46, 3.33]} & {\scriptsize[$-$0.99, 0.70]} \\
        $z$-gate $-$ Magnitude-cap & $+0.20$ & $+2.03$ & $-0.55$ & $+0.78$ & $+3.13$ & $+1.04$ & $+1.10$ \\
        & {\scriptsize[$-$1.55, 1.95]} & {\scriptsize[$-$0.31, 4.69]} & {\scriptsize[$-$2.57, 1.38]} & {\scriptsize[$-$0.81, 2.41]} & {\scriptsize[0.42, 6.46]} & {\scriptsize[$-$1.04, 4.17]} & {\scriptsize[0.20, 2.07]} \\
        Top-$\alpha$ $-$ Magnitude-cap & $+0.50$ & $+1.41$ & $-0.46$ & $+1.85$ & $+3.54$ & $-0.21$ & $+1.11$ \\
        & {\scriptsize[$-$1.15, 2.20]} & {\scriptsize[$-$1.72, 4.22]} & {\scriptsize[$-$2.67, 1.75]} & {\scriptsize[0.30, 3.44]} & {\scriptsize[0.21, 7.50]} & {\scriptsize[$-$1.67, 1.04]} & {\scriptsize[0.13, 2.12]} \\
        $z$-gate $-$ Top-$\alpha$ & $-0.30$ & $+0.63$ & $-0.09$ & $-1.07$ & $-0.42$ & $+1.25$ & $-0.00$ \\
        & {\scriptsize[$-$2.00, 1.35]} & {\scriptsize[$-$2.19, 3.59]} & {\scriptsize[$-$2.30, 2.02]} & {\scriptsize[$-$2.56, 0.41]} & {\scriptsize[$-$3.75, 2.50]} & {\scriptsize[$-$1.04, 4.38]} & {\scriptsize[$-$1.01, 1.01]} \\
        \bottomrule
    \end{tabular}
    \caption{Paired differences in percentage points among our run-1 models, with 95\% paired bootstrap intervals.}
    \label{tab:supp-paired}
\end{table*}

\begin{table*}[tb]
    \centering
    \small
    \setlength{\tabcolsep}{5pt}
    \begin{tabular}{lcccccc}
        \toprule
        Model & Clipped steps & Resp.\ length & Student entropy & Weighted & Unweighted & Mean \\
        & (1--50 / 51--200) & (first / last 50) & (first / last 50) & bound.\ (\%) & resp.\ (\%) & weight \\
        \midrule
        Vanilla OPD & 43 (36 / 7) & 1464 / 2437 & 0.764 / 0.168 & -- & -- & 1 \\
        AP-OPD (top-$\alpha$) & 30 (30 / 0) & 1441 / 2432 & 0.757 / 0.174 & 6.52 & 1.57 & 0.952 \\
        AP-OPD ($z$-gate) & 33 (33 / 0) & 1485 / 2408 & 0.816 / 0.173 & 6.79 & 24.22 & 0.945 \\
        Magnitude-cap & 32 (31 / 1) & 1440 / 2434 & 0.726 / 0.175 & 6.52 & 1.33 & 0.961 \\
        \bottomrule
    \end{tabular}
    \caption{Run-1 training dynamics over 200 steps, reported as means of logged values.
    Clipped steps have pre-clipping gradient norm above 1.}
    \label{tab:supp-dynamics}
\end{table*}

\section{Alternative Reweighting Rules}
\label{sec:supp-negative}
AP-OPD uniformly scales supervision in the post-jump block using the preceding block's original block-averaged penalty as a local reference.
Two alternatives are natural: compensating by upweighting the pre-jump block and down-weighting the post-jump block by a fixed constant.
We evaluate three such rules at step 100 with 8{,}192-token evaluation responses, so their numbers are not comparable with Table~\ref{tab:main}.
All three select boundaries with the top-$\alpha$ ranking of Eq.~\eqref{eq:topk} with $\alpha=5\%$.
The paired rule raises each selected pre-jump block and lowers its post-jump block by the same amount, with the weights normalized over the response.
The two one-sided rules multiply every selected pre-jump block by 4 or every selected post-jump block by 0.3.

\begin{table*}[!tbp]
    \centering
    \small
    \setlength{\tabcolsep}{3pt}
    \begin{tabular}{lccccccccc}
        \toprule
        Rule & MATH500 & AMC23 & Minerva & Olympiad & AIME24 & AIME25 & Macro & Clipped & Mean $w$ \\
        \midrule
        Vanilla OPD & 69.90 & 43.12 & 20.31 & 30.67 & 11.46 & 6.25 & 30.29 & -- & 1 \\
        Paired ($+$pre, $-$post) & 70.00 & 41.88 & 20.77 & 30.33 & 11.67 & 7.71 & 30.39 & 32 / 100 & 1.006 \\
        Pre-jump $\times4$ & 69.15 & 41.88 & 20.50 & 31.26 & 12.29 & 6.25 & 30.22 & 88 / 100 & 1.200 \\
        Post-jump $\times0.3$ & 70.25 & 44.38 & 21.23 & 31.11 & 12.29 & 6.88 & 31.02 & 30 / 100 & 0.955 \\
        \bottomrule
    \end{tabular}
    \caption{Alternative reweighting rules at step 100, reported as Avg@$k$ in percent from one training run with 8K evaluation responses.
    Clipped: steps whose pre-clipping gradient norm exceeds 1; Mean $w$: mean token weight.}
    \label{tab:supp-negative}
\end{table*}

Table~\ref{tab:supp-negative} shows that upweighting the pre-jump block brings no gain.
The pre-jump $\times4$ rule triggers gradient clipping in 88 of 100 steps.
The paired rule has a macro average close to vanilla OPD.
Down-weighting the post-jump block by a constant 0.3, fixed before training, improves over vanilla OPD and has the best macro average.
AP-OPD uses the preceding block's original mean to determine the post-jump weight, $D_i/D_{i+1}$, without an additional scaling constant.
The fixed-weight result shows that a gain can also occur under another post-jump weighting rule; it does not establish that the ratio is superior, since the protocols differ.

\section{Block-Level versus Token-Level Down-Weighting}
\label{sec:supp-reopold}
This appendix gives the protocol and additional numbers for \S\ref{sec:analysis}.

\paragraph{Setup.}
REOPOLD bounds each token's penalty from above \citep{ko2026reopold}; in our notation it replaces $r_t$ by $\min(r_t,\kappa)$ with
\begin{equation}
\kappa=\frac{-\log\lambda}{1-\lambda}.
\end{equation}
We use two bounds.
In the removal-matched setting, $\lambda=0.051$ gives $\kappa=3.131$, which removes the same 11.6\% of penalty mass as AP-OPD on the analysis set.
Under REOPOLD's default, $\lambda=0.3$ gives $\kappa=1.720$, which removes 27.0\%.
On the two 300-step runs of Appendix~\ref{sec:supp-diagnosis} we keep both bounds fixed rather than recalibrating them, so the matched bound removes somewhat more than AP-OPD there: 13.2\% and 13.4\%, compared with 10.7\% and 10.6\% for AP-OPD, respectively.
After clipping, we recompute the block profile, rerun the $z=2$ detector, and repeat the split-half measurement of Appendix~\ref{sec:supp-splithalf}.

\begin{table*}[!tbp]
    \centering
    \small
    \setlength{\tabcolsep}{6pt}
    \begin{tabular}{llccccc}
        \toprule
        Rollouts & Bound & Removed & Jumps kept & Rise kept & Overlap & Expected \\
        \midrule
        Analysis set & matched & 11.6 & 99.0 & 89.8 {\scriptsize[86.1, 93.8]} & 14.5 {\scriptsize[13.1, 15.9]} & 12.5 \\
        Analysis set & default & 27.0 & 96.3 & 71.1 {\scriptsize[67.0, 75.8]} & 32.3 {\scriptsize[30.9, 33.6]} & 28.6 \\
        \midrule
        Vanilla, 300 steps & matched & 13.2 & 98.9 & 86.1 {\scriptsize[84.8, 87.4]} & 13.3 {\scriptsize[13.1, 13.6]} & 13.1 \\
        Vanilla, 300 steps & default & 30.1 & 95.9 & 68.0 {\scriptsize[66.7, 69.4]} & 31.9 {\scriptsize[31.6, 32.2]} & 30.2 \\
        \midrule
        Paired rule, 300 steps & matched & 13.4 & 98.7 & 84.2 {\scriptsize[82.9, 85.5]} & 13.7 {\scriptsize[13.5, 14.0]} & 13.1 \\
        Paired rule, 300 steps & default & 30.3 & 95.5 & 66.4 {\scriptsize[65.0, 67.8]} & 32.5 {\scriptsize[32.1, 32.8]} & 30.3 \\
        \bottomrule
    \end{tabular}
    \caption{Token-level clipping on the analysis set and the two 300-step runs at $B=64$ and $z=2$.
    Values are percentages; intervals are 95\%.
    Matched: $\kappa=3.131$, which removes as much penalty mass as AP-OPD on the analysis set; default: REOPOLD's $\kappa=1.720$.
    Removed: penalty mass removed by clipping, averaged over responses.
    Jumps kept: original jumps detected again after clipping.
    Rise kept: held-out rise $U_{0:64}$ after clipping relative to no clipping.
    Overlap: share of AP-OPD's removal that lies above the bound.
    Expected: share of all penalty mass above the bound, the overlap expected if AP-OPD's removal were unrelated to token size.
    Overlap and Expected pool mass over tokens.}
    \label{tab:supp-clip}
\end{table*}

\paragraph{Rise after clipping.}
Table~\ref{tab:supp-clip} shows that at both bounds and on all three sets of rollouts, nearly all jumps are detected again and most of the held-out rise remains.
At the matched bound on the analysis set, the held-out rise still exceeds each of the three matched controls in both detection directions under the paired bootstrap of Appendix~\ref{sec:supp-splithalf}, as it does at the default bound.
Clipping also changes block variances and creates new jumps.
The detector finds 1{,}124 jumps at the matched bound and 1{,}296 at the default bound, compared with 978 without clipping.
The retained rise therefore compares two detected sets rather than following each original jump.

\paragraph{Overlap of removed mass.}
With AP-OPD's selections and weights fixed on the unclipped signal, we split each removed amount $(1-w_t)d_t$ into the part above $\kappa$, which clipping would also remove, and the part below it.
If AP-OPD's removal were unrelated to token size, the part above $\kappa$ would equal the share of all penalty mass that lies above $\kappa$.
Table~\ref{tab:supp-clip} shows that the overlap stays close to this reference at both bounds and on all three sets of rollouts.
On the two 300-step runs at the matched bound, it nearly coincides with the reference. The two rules therefore remove largely different supervision.

\end{document}